\documentclass[runningheads]{llncs}

\usepackage{graphicx}
\usepackage{algorithm}
\usepackage{algpseudocode}
\usepackage{outlines}
\usepackage{bm}
\usepackage{amsmath}
\DeclareMathOperator*{\argmax}{arg\,max}

\usepackage{amssymb}
\usepackage{wrapfig}
\usepackage{tikz}
\usepackage{pgfplots}
\usepackage{pgfplotstable}
\usepgfplotslibrary{groupplots}
\usepackage{array}
\usepackage{siunitx}
\usepackage{booktabs}
\usepackage{multirow}
\usepackage{cleveref}
\usepackage{filecontents}
\usepackage{wrapfig}

\pgfplotsset{compat=1.16}

\begin{document}
\title{Informative Communication of Robot Plans}

\author{Michele Persiani,
Thomas Hellstr\"om\\ Ume{\aa} University, Ume{\aa}, Sweden\\
\textit{michelep@cs.umu.se,thomas.hellstrom@umu.se}}

\author{Michele Persiani\inst{1}\orcidID{0000-0001-5993-3292} \and
Thomas Hellstr\"om\inst{1}\orcidID{0000-0001-7242-2200}}
\authorrunning{Michele Persiani and Thomas Hellstr\"om}
%
\institute{Ume{\aa} University, Ume{\aa}, Sweden\\
\email{michelep@cs.umu.se, thomas.hellstrom@umu.se}}

\maketitle

\begin{abstract}

When a robot is asked to verbalize its plan it can do it in many ways. For example, a seemingly natural strategy is incremental, where the robot verbalizes its planned actions in plan order. However, an important aspect of this type of strategy is that it misses considerations on what is effectively informative to communicate, because not considering what the user knows prior to explanations. In this paper we propose a verbalization strategy to communicate robot plans informatively, by measuring the information gain that verbalizations have against a second-order theory of mind of the user capturing his prior knowledge on the robot. As shown in our experiments, this strategy allows to understand the robot's goal much quicker than by using strategies such as increasing or decreasing plan order. In addition, following our formulation we hint to what is informative and why when a robot communicates its plan.

\end{abstract}


\section{Introduction}
\label{sec:introduction}

With its ever-growing advancements, Artificial Intelligence (AI) is proving to be a promising partner in our lives. In envisioning this relationship, AI agents should support us by being able to perform their assigned tasks both efficiently and accurately, but, on top of that, a crucial aspect for this relationship to be successful is the degree by which agents are able to make themselves understood by their human users, either through explanations, or by behavior that is interpretable from the human perspective \cite{hellstrom2018understandable}.

An interpretable behavior is a behavior that is expected by the user, because fulfilling his expectations about the agent \cite{chakraborti2017plan,chakraborti2019explicability}. Once implemented in a model, these expectations describes how the user is modeling the agent, therefore informing the agent about how it is being perceived and explained. It follows that behaviors fitting this expectations model are intrinsically explainable and therefore should not require additional explanations; however, other types behaviors could require explanations. This is because, while fulfilling other relevant properties such as being optimal, do not fit the user's expectations, and thus result being uninterpretable. It becomes therefore crucial to complement such behavior with explanations targeting its alignment with the user's expectation model, such that it becomes interpretable in it \cite{chakraborti2017plan}.

\begin{wrapfigure}{r}{0.5\textwidth}
    \centering
    \includegraphics[width=\linewidth]{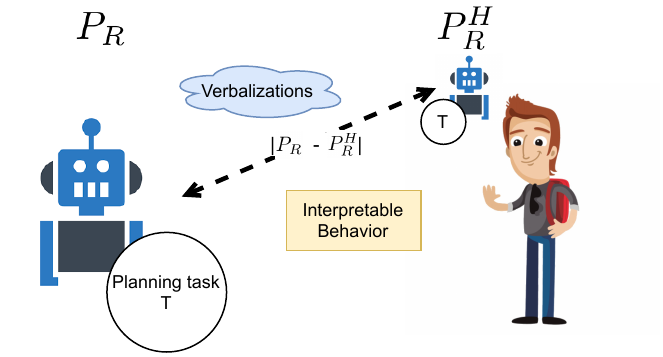}
    \caption{\footnotesize $P_R$: model of the robot working on task $T$. $P_R^H$: second-order theory of mind of the robot. The goal of interpretable behavior and verbalizations is to reduce the distance $|P_R - P_R^H|$. Here we consider the minimum amount of actions, and their order, that the robot should communicate such that $|P_R - P_R^H| < \theta$.}
    \label{fig:front}
\end{wrapfigure}

A relevant measure of the explanation process is its degree of \textit{informativeness}, that is the amount of information that explanations transfer to the explainee \cite{lawless2019artificial}. To be informative, explanations can leverage the user's expectation model on the agent when computing what to communicate and how: given that many candidate verbalizations are possible to communicate, the agent has to determine which one is most informative, in the sense that it increases by a largest amount the user's understanding of the agent. We see the process of explanation as having a complementary role with respect to interpretable behavior, and explanations and interpretable behavior have the same objective of keeping the agent model $P_R$, and the expectations that the user has about the agent $P_R^H$, as similar as possible. This is because, when aligned, this alignment signals that the agent is being understood because fitting the expectations.

Informativeness in explanations is still sparsely considered inside the literature, and previous work on explanations focuses more on the consistency of explanations from the agent’s perspective, rather than generating explanations that are informative to the user \cite{gavriilidis2021plan,moon2019towards}. With the goal of addressing this gap, in this paper we propose an algorithm to communicate a robot's plan using a criteria based on the informativeness of communications. We achieve this by firstly grounding the understanding that the user has about the agent as a measure of difference between agent model and user expectation model. Later, we select which actions of the plan to communicate, and their order, based on how their communication affect this distance measure. In addition, by the same concepts we define what is informative in the case of explanations of robot plans.

The rest of the paper is structured as follows. In Section~\ref{sec:background} we provide brief, relevant background in theory of mind for robotics and verbalizations of robot plans. In Section~\ref{sec:method} we present our method for informative communication of the robot plan. Section~\ref{sec:experiments} describes experiments and measures performed on different tests. Finally, Section~\ref{sec:discussion} and Section~\ref{sec:conclusion} contains discussion on the results and conclusions.

\section{Background}\label{sec:background}

Previous research in understandable robots \cite{hellstrom2018understandable} propose to consider the concept of \textit{understanding} as a distance measure between a robot's intentional model $P_R$ and the user's model of $P_R$, $P_R^H$. The more similar these models are, the higher is the degree of understanding between user and robot. The limit case where the distance is $0$ signals the robot that it is being fully understood in the sense that the user's expectations of it are the same as its actual intentional model, while a measure $>0$ signals the robot that some parts of its intentional model are mismatching with the user's expectation, and therefore require additional verbalizations. Here we see verbalizations as sequences of communicative actions\footnote{A communicative action is an action, either verbal or physical, intended to decrease the models distance $|P_R - P_R^H|$ \cite{hellstrom2018understandable}} aimed at communicating parts of the intentional model. While having the same function of interpretable behavior, that is to reduce $|P_R - P_R^H|$, they are purely communicative in nature, such as through speech, text, images, etc.

The model $P_R^H$ constitutes a second-order theory of mind \cite{hellstrom2018understandable}, and shown in recent literature, can take many forms. For example, in \cite{zhang2017plan} and in \cite{gong2018robot} it is a context-dependent label describing whether the user understands the robot. After a training phase through human annotations, predicting the label's value allows to augment the cost function of the planning procedure with explainability awareness. Alternatively, more structured forms are for example presented in \cite{chakraborti2017plan}, where the theory of mind is a complete planning instance.

The work on understandable robots has connections also to research in interpretability and explainability. Given that a model of the user's expectations $P_R^H$ is provided to the robot (Fig.~\ref{fig:front}), interpretable behaviors keep the distance measure $|P_R - P_R^H|$ low by modifying its behavior, while explanations through sequences of communicative actions (that we refer to as verbalizations) \cite{kambhampati2019synthesizing,sreedharan2017balancing}.

In this paper we focus on explanations, and in particular on the verbalizations of the robot's task experience, such as its recent course of action, its goal, or plan. For this type of question, the strategies to verbalize explanations that are proposed in the literature are often incremental i.e. orderly from the first to the last actions. For example, \cite{rosenthal2016verbalization,meza2016towards} propose strategies to verbalize robot plans. In addition, the authors keep into consideration possible categories of users by predefining, for a given plan, multiple types of explanations, one for every type of user. The resulting explanations span over standardized dimensions of verbalization such as abstraction, locality, and specificity. However, while these strategies can consider different types of users and what can be informative to them, they don't consider the information that users (in any given category) possess prior to the explanations, and therefore verbalizations can only be incremental, or with hand-coded strategies. \cite{zhu2017autonomous} addresses this problem by a manual approach: we can ask the users which elements of the plan are of interest before its verbalization, in this way, the successive incremental verbalization can be filtered to contain only elements that are relevant to the specific user.

As we show in this paper the order by which actions are verbalized can be controlled to maximize the amount of information that verbalizations produce inside the second-order theory of mind. A similar idea is explored in \cite{zakershahrak2020order}, where the authors propose how the explanation process can be re-ordered by keeping into consideration the cognitive load of the user. While we also confront the question of ordering verbalizations, in this paper we are interested in making the user understand the robot's internal state with a minimum amount of utterances.

\section{Method}
\label{sec:method}

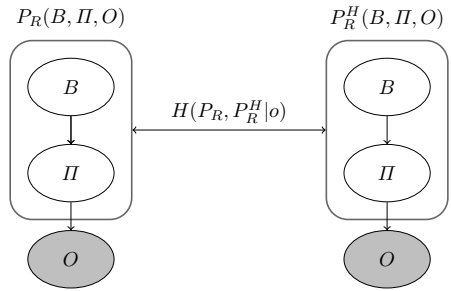
\begin{wrapfigure}{l}{0pt}
    \resizebox{.5\linewidth}{!}{
    \begin{tikzpicture}[align=center] 
    \usepgflibrary{shapes.geometric}
    \usetikzlibrary{shapes,backgrounds,shapes.misc, positioning,shapes.geometric,arrows,matrix,fit,calc}
    \tikzstyle{main}=[draw, ellipse, minimum height=1cm, minimum width=1.5cm, align=center]

    \node[main]                         at (0, 0)    (1) {$B$};
    \node[main]                         at (0, -1.5) (2) {$\Pi$};
    \node[main, fill=lightgray]         at (0, -3)   (3) {$O$};
    \node[draw=white,text width=2cm]    at (0,1.2) {$P_R(B,\Pi,O)$};
    
    \node[main]                         at (5.5, 0)    (4) {$B$};
    \node[main]                         at (5.5, -1.5) (5) {$\Pi$};
    \node[main, fill=lightgray]         at (5.5, -3)   (6) {$O$};
    \node[draw=white,text width=2cm]    at (5.5,1.2) {$P_R^H(B,\Pi,O)$};
    
    \begin{scope}[on background layer]
        \node[draw, thick, rounded corners = 2.5ex, fit=(1) (2),inner sep=3mm, opacity=0.6](Fit1) {};
        \node[draw, thick, rounded corners = 2.5ex, fit=(4) (5),inner sep=3mm, opacity=0.6](Fit2) {};
    \end{scope}

    \draw[->] (1) -- (2);
    \draw[->] (2) -- (3);
    \draw[->] (4) -- (5);
    \draw[->] (5) -- (6);

    \draw[<->] (Fit1) -- node [midway,above ] {$H(P_R, P_R^H|o)$} (Fit2);
    
        \draw[->] (1) -- (2);

    \end{tikzpicture}
    }
    \caption{\footnotesize Robot model ($P_R$) and second-order theory of mind of the user ($P_R^H$) as equivalent Bayesian Networks. The cross-entropy $H$ measures the difference between robot's and user's estimated state of mind a posteriori of a verbalization $o$. }
    \label{fig:model}
\end{wrapfigure}

We consider the case in which the robot's model $P_R$  and user's estimated model of the robot model, $P_R^H$, are probabilistic Belief-Desire-Intention (BDI) models, described by two Equivalent Bayesian Networks and forming a second-order theory of mind setting (Figure~\ref{fig:model}). The networks use the same random variables, however, these variables can be differently distributed in $P_R$ and $P_R^H$, thus possibly reflecting a mismatch between the user's beliefs over the robot model and the true robot model. We refer to this setting as the \textit{Mirror Agent Model} because defining the second-order theory of mind of the agent as a ``mirror'' of the agent's true model. This model is easy to deploy and can be obtained \textit{as a service} \cite{cashmore2019towards} on a variety of agents, which is an important property for its usability.

A setting which mirrors $P_R$ and $P_R^H$ has clear advantages when the robot decides what to communicate in order to increase the models similarity: if we are able to directly compare the robot and its theory of mind, we could immediately know what part of the model is being understood, and which part instead requires a verbalizations. However, the assumption this setting makes is that the user has internalized a robot model of the same form of the original. For this paper we take this assumption to hold, and model the user as knowing the exact robot model\footnote{This assumption doesn't make the proposed method to lose generality, because from two different $P_R$ and $P_R^H$ we can always create a super-agent which comprehend both models, and that is used in the mirror model. Then, we would set inside $P_R$ and $P_R^H$ the corresponding probabilities to 0.}. The robot's probabilistic BDI model is defined $P_R(O,\Pi,B) = P_R(O|\Pi)P_R(\Pi|B)P_R(B)$ where every $b\in B$ include both belief and goal of the robot, $\pi \in \Pi$ a possible intention (we consider intentions as plans) and $o\in O$ the observations that plans yield to the user. Since we here consider the case where the robot verbalizes its plans, observations are verbalizations of actions, that in our case have the form of textual descriptions of sequences of actions. The only observable random variable is $O$, while $B$ and $\Pi$ are variables internal to the robot's \textit{state of mind} \cite{hellstrom2018understandable}, and can be interacted with only through observations. $P_R^H$ is similarly defined. 

As a simplified case, we don't consider how the robot forms and updates its beliefs through its sensors, and we assume a fixed probabilistic belief $P_R(B)$. The robot can form plans using the planning model $P_R(\Pi|B)$, where the probability of a plan $P_R(\pi|b)$ is $>0$ only if the plan is consistent with both belief and goal condition of $b$. While computing plans we model the robot to be a rational agent ie. associating higher probabilities to cheap plans, and accordingly set $P_R(\pi|b) \propto R(\pi, b)$ where $R(.)$ is a function measuring the rationality of $\pi$ when executed in $b$ \cite{persiani2021inference}. The model $P_R(O|\Pi)$ instead describes how executing plans create observations. $P_R(o|\pi) > 0$ means that $o$ verbalizes a set of actions $o=\{a_1, ...,a_n\}$ that are a subset of the plan $\pi$, with $P_R(o|\pi) = \prod_i P_R(a_i|\pi), P_R(a_i|\pi) > 0 \: \mathbf{iff}\: a_i \in \pi$. The model $P_R^H$ is similarly defined. The cross-entropy measure $H(P_R, P_R^H)$ determines the similarity between $P_R$'s and $P_R^H$'s states of mind (ie. their corresponding intentions and beliefs). For simplicity, we consider the case of a deterministic $P_R$, with $P_R(B=b_R)=1$ and $P_R(\Pi=\pi_R)=1$, and probabilistic $P_R^H$. In this setting, the cross-entropy between $P_R$ and $P_R^H$ is:
\begin{align}
    H(P_R,P_R^H) = \sum P_R(\pi, b) \log P_R^H(\pi, b)= -\log P_R^H(\pi_R|b_R) - \log P_R^H(b_R)
\end{align}

Therefore, to increase the similarity between $P_R$ and $P_R^H$ we must increase the probability of $\pi_R$ and $b_R$ in the theory of mind; the limit case where also the theory of mind deterministically produces $\pi_R$ and $b_R$ is the case in which the user is estimated to have an exact understanding of the robot. In this case the cross-entropy is 0.

\subsection{Informative Communication}

As described previously, the main objective of a verbalization $o\in O$ is to make $P_R^H$ as similar as possible to $P_R$. This is achieved by decreasing the cross-entropy measure between the models. The associated Information Gain (IG) of the verbalization is $ IG(P_R,P_R^H,o) = H(P_R, P_R^H) - H(P_R,P_R^H|o) $. Using the information gain, the most informative verbalization can be selected as the one with maximal information gain:
\begin{align}
     \hat{o} = &\argmax_{o\in O}\:  H(P_R, P_R^H) -  H(P_R,P_R^H|o)\nonumber \\
             = &\argmax_{o\in O}\: \log P_R^H(\pi_R,b_R|o)\nonumber \\
             = &\argmax_{o\in O}\: \log P_R^H(o|\pi_R) - \log \mathbb{E}[ P_R^H(o|\pi)] \label{eq:best_o_e}
\end{align}

Two important properties of informative communication emerge from Eq.~\ref{eq:best_o_e}. The first is that only verbalizations that correctly reflect the model of the robot are informative. Communicating an action not belonging to the robot's plan, with $P_R^H(a_i|\pi_R) = 0$, induce an informativeness of $-\infty$ of the full verbalization. The second is that actions that are informative to communicate are those that tend to appear exclusively in $\pi_R$. 

A simple way to find which is the best verbalization of $N$ actions to communicate is to enumerate all the possible combinations of actions belonging to $\pi_R$, to then select the combination with highest IG. Algorithm~\ref{alg:inf_comm} implements this procedure. The output of the algorithm is the sequence of the $N$ most informative actions ordered in plan order.

\begin{algorithm}[H]
  \caption{Find the most informative communication of size $N$, by enumerating and sorting the combinations of actions of size $N$}
  \label{alg:inf_comm}
  \begin{algorithmic}[1]
  \Procedure{Verbalize-Plan}{$P_R, P_R^H, N$}
  \State $\pi_R \gets \Call{Plan}{P_R}$
  \State $o \gets $ \Call{Find-Most-Informative}{$\pi_R, P_R^H, N$}
  \State $o_{sorted} \gets $ \Call{Plan-Sort}{$o$}
  \State \Call{Verbalize}{$o_{sorted}$}
  \EndProcedure
  \end{algorithmic}
  \begin{algorithmic}[1]
  \Procedure{Find-Most-Informative}{$\pi_R, P_R^H, N$}
  \State $C \gets \Call{Combinations}{\pi_R, N}$ \Comment{$|C|=\binom{|\pi_R|}{N}$ }
  \State $Q\gets \emptyset$
  \For{$c\in C$}
    \State $h_c = $ \Call{Information-Gain}{$P_R^H|o=c$}  \Comment{Eq.~\ref{eq:best_o_e}}
    \State \Call{Append}{$Q, \langle h_c,c \rangle$}
  \EndFor
  \State $Q \gets$ \Call{Sort}{$Q$}
  \State $h_{best}, c_{best} \gets$ \Call{Pop}{$Q$}
  \State $\mathbf{yield}\:c_{best}$
  \EndProcedure
  \end{algorithmic}
\end{algorithm}

\section{Implementation in PDDL}
\label{sec:pddl}

We implemented the probabilistic BDI models $P_R(O,\Pi,B)$ and $P_R^H(O,\Pi,B)$ by specifying planning instances using the Planning Domain Description Language (PDDL). PDDL \cite{McDermott1998PDDLthePD} is a standard language to specify planning domains for what is usually referred to as classical planning. A planning instance for the robot is obtained by specifying the tuple $\langle \mathcal{P}_R, \mathcal{A}_R, I_R, \mathcal{G}_R, \mathcal{O}_R\rangle$. Where $I_R$ and $\mathcal{G}_R$ are set of ground predicates and correspond to the initial and goal state respectively, $\mathcal{O}_R$ is the set of objects available to ground the predicates $\mathcal{P}_R$, while $\mathcal{A}_R$ is the set of available actions to transition between states. Similarly, the second order theory of mind model has components $\langle \mathcal{P}_R^H, \mathcal{A}_R^H, I_R^H, \mathcal{G}_R^H, \mathcal{O}_R^H\rangle$.  

We set the descriptive components of the planning instances of robot and theory of mind to be equivalent. ie. $\mathcal{P}_R = \mathcal{P}_R^H, \mathcal{A}_R = \mathcal{A}_R^H$ and $\mathcal{O}_R = \mathcal{O}_R^H$, with the only probabilistic parts being $I_R$, $\mathcal{G}_R$, $I_R^H$ and $\mathcal{G}_R^H$. In this setting, the corresponding probability distribution over the possible PDDL instances describing the robot state is obtaned by a combination of Bernoulli distribution for the beliefs $I_R$, and a categorical distribution for the possible goals $\mathcal{G}_R$ (the same for $I_R^H$ and $\mathcal{G}_R^H$ respectively).

\begin{align*}
&P_R(B) = P_R(I)P_R(G)                                   &      &P_R(G;\theta_R) = P(G|\{g_0,...,g_m\})   \\
&P_R(I;\theta_R) = \Pi_i P(p_{i} \in I_R;\theta_{p_i})   &      &P_R(G= \langle \mathcal{G}_j \rangle|\{g_0,...,g_m\}) = \theta_j\\
&P(p_{i} \in I_R) = \theta_i                             &       &\sum_j \theta_j = 1
\end{align*}

where $\mathbf{\theta}$ are the distribution parameters. Sampling a belief from $P_R(B)$ yields a initial state and a goal state for the PDDL planner. The planning model $P_R(\Pi|B)$ is instead implemented by a planner of choice compatible with the underlying PDDL requirements. The probability of a plan $P_R(\Pi=\pi|B=b)$ is defined as a function of rationality \cite{persiani2021inference}. 
Sampling from the planning model can for example be done through Diverse Planning techniques \cite{katz2020reshaping}. $P_R^H(B)$ and $P_R^H(\Pi|B)$ are similarly defined.

\section{Experiments}
\label{sec:experiments}
We test our algorithm for informative communication over two sets of tests. First, we run a set of automated benchmarks on the PUCRS dataset, which is a curated dataset of planning domains \cite{pereira2017landmark}, discussing then some generally valid metrics and results. After that we show the results of a user study simulating the case where an operator meets a robot working in a warehouse, and not knowing what it is doing, asks for its plan.

\subsection{Tests on the PUCRS dataset}

We benchmarked Algorithm~\ref{alg:inf_comm} on the PUCRS dataset \cite{pereira2017landmark}, that is a dataset of PDDL planning domains and problems, with the goal to search for properties of informative communication that generalizes across domains. We selected 6 standard PDDL domain: \textit{logistics, intrusion-detection, rovers, satellite, blocks-world, satellite}. Table~\ref{tbl:pucrs_dataset} shows relevant average measures of the domains, such as number of predicates and operators, size of the initial state, number of tested goals and length of optimal plans.

For every domain we perform the tests on 10 planning instances, averaging the results. In each test $P_R$ is initialized by randomly selecting a goal from the pool of available goals and using the original instance's initial state. $P_R^H$ is initialized by randomly selecting 3 goals from the goal pool plus the goal of $P_R$, for a total of 4 possible goals. These goals are used with equal probabilities in $P_R^H(G)$. In addition, 4 random predicates belonging to the robot plan's preconditions are made as probabilistic ie. $P_R^H(I)=\underset{i\in 1..4}{\prod} P(p_i)$, with $P(p_i=1)=0.5$.

\begin{wraptable}{r}{.5\linewidth}
\centering
  \begin{tabular}{|c|c|c|c|c|c|}
    \hline
    \textbf{Domain} & $|\mathcal{P}|$  & $|\mathcal{A}|$  &  $|I|$ & $|G|$ & $|\overline{\pi}|$\\ 
    \hline
    \textit{miconic}	            & 8	    &	4	&	517	    &	6       &	35.57 \\
    \textit{logistics}              & 3	    &	6	&	22.16	&	10.39	&	24.41 \\
    \textit{intrusion}           	& 11	&	9	&	1	    &	16.67	&	13.07 \\
    \textit{blocks-world}        	& 5	    &	4	&	14.58	&	20.28	&	14.50 \\
    \textit{rovers}              	& 32	&	9	&	172.71	&	6       &	24.93 \\
    \textit{satellite}            	& 12	&	5	&	58	    &	6.43	&	16.89 \\
    \hline
  \end{tabular}
    \\[2mm]
  \caption{Average instance measures over the tested planning domains. The columns, from left to right are: number of operators, number of predicates, size of the initial state, number of tested goals, length of optimal plans.}
 \label{tbl:pucrs_dataset}
\end{wraptable}

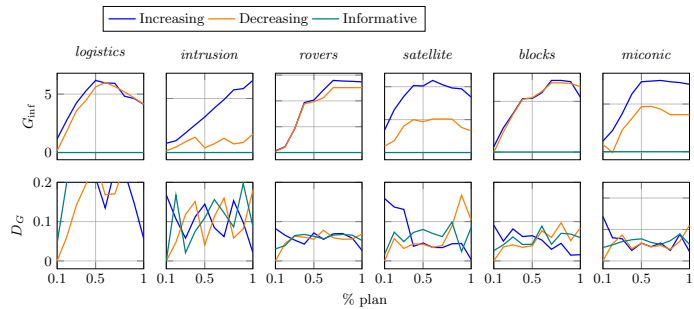
\begin{figure*}
    \centering
    \scalebox{.6}{
    \begin{tikzpicture}
    \begin{groupplot}[group style={
                              group size=6 by 2,
                              group name=plots,
                              vertical sep=0.5cm,
                              horizontal sep = 0.5cm,
                              ylabels at=edge left,
                              xlabels at=edge bottom,
                              xticklabels at=edge bottom},
                              width=3.5cm,
                              height=3.5cm,
                              try min ticks=3,
                              xtick={0.1, 0.5, 1},
                              xmax=1,
                              xmin=.1,
                              ]
                           
              

    \nextgroupplot[title = {\emph{logistics}}, grid=both, every major grid/.style={gray, opacity=0.5}, ylabel={$G_{\text{inf}}$}, legend style={legend columns=3},legend to name={CommonLegend}]
    \addplot[thick,blue] table [x=x, y=e_gain_inc, col sep=comma] {data/pucrs/logistics.csv};
    \addplot[thick,orange] table [x=x, y=e_gain_dec, col sep=comma] {data/pucrs/logistics.csv};
    \addplot[thick,teal] table [x=x, y=e_gain_ent, col sep=comma] {data/pucrs/logistics.csv};
    \addlegendimage{orange, mark=*}
    \addlegendentry{Increasing}
    \addlegendentry{Decreasing}
    \addlegendentry{Informative}

    \nextgroupplot[title = {\emph{intrusion}}, grid=both, every major grid/.style={gray, opacity=0.5},yticklabels={,,}]
    \addplot[thick,blue] table [x=x, y=e_gain_inc, col sep=comma] {data/pucrs/intrusion-detection.csv};
    \addplot[thick,orange] table [x=x, y=e_gain_dec, col sep=comma] {data/pucrs/intrusion-detection.csv};
    \addplot[thick,teal] table [x=x, y=e_gain_ent, col sep=comma] {data/pucrs/intrusion-detection.csv};
                              
    \nextgroupplot[title = {\emph{rovers}}, grid=both, every major grid/.style={gray, opacity=0.5}, yticklabels={,,}]
    \addplot[thick,blue] table [x=x, y=e_gain_inc, col sep=comma] {data/pucrs/rovers.csv};
    \addplot[thick,orange] table [x=x, y=e_gain_dec, col sep=comma] {data/pucrs/rovers.csv};
    \addplot[thick,teal] table [x=x, y=e_gain_ent, col sep=comma] {data/pucrs/rovers.csv};
                              
    \nextgroupplot[title = {\emph{satellite}}, grid=both, every major grid/.style={gray, opacity=0.5},yticklabels={,,}]
    \addplot[thick,blue] table [x=x, y=e_gain_inc, col sep=comma] {data/pucrs/satellite.csv};
    \addplot[thick,orange] table [x=x, y=e_gain_dec, col sep=comma] {data/pucrs/satellite.csv};
    \addplot[thick,teal] table [x=x, y=e_gain_ent, col sep=comma] {data/pucrs/satellite.csv};
                              
    \nextgroupplot[title = {\emph{blocks}}, grid=both, every major grid/.style={gray, opacity=0.5},yticklabels={,,}]
    \addplot[thick,blue] table [x=x, y=e_gain_inc, col sep=comma] {data/pucrs/blocks-world.csv};
    \addplot[thick,orange] table [x=x, y=e_gain_dec, col sep=comma] {data/pucrs/blocks-world.csv};
    \addplot[thick,teal] table [x=x, y=e_gain_ent, col sep=comma] {data/pucrs/blocks-world.csv};
                              
    \nextgroupplot[title = {\emph{miconic}}, grid=both, every major grid/.style={gray, opacity=0.5},yticklabels={,,}]
    \addplot[thick,blue] table [x=x, y=e_gain_inc, col sep=comma] {data/pucrs/miconic.csv};
    \addplot[thick,orange] table [x=x, y=e_gain_dec, col sep=comma] {data/pucrs/miconic.csv};
    \addplot[thick,teal] table [x=x, y=e_gain_ent, col sep=comma] {data/pucrs/miconic.csv};


    \nextgroupplot[grid=both, every major grid/.style={gray, opacity=0.5}, ylabel={$D_G$}, ymax=0.2]
    \addplot[thick,blue] table [x=x, y=g_dist_inc, col sep=comma] {data/pucrs/logistics.csv};
    \addplot[thick,orange] table [x=x, y=g_dist_dec, col sep=comma] {data/pucrs/logistics.csv};
    \addplot[thick,teal] table [x=x, y=g_dist_ent, col sep=comma] {data/pucrs/logistics.csv};
                              
    \nextgroupplot[grid=both, every major grid/.style={gray, opacity=0.5}, yticklabels={,,}, ymax=0.2]
    \addplot[thick,blue] table [x=x, y=g_dist_inc, col sep=comma] {data/pucrs/intrusion-detection.csv};
    \addplot[thick,orange] table [x=x, y=g_dist_dec, col sep=comma] {data/pucrs/intrusion-detection.csv};
    \addplot[thick,teal] table [x=x, y=g_dist_ent, col sep=comma] {data/pucrs/intrusion-detection.csv};
                              
    \nextgroupplot[grid=both, every major grid/.style={gray, opacity=0.5}, yticklabels={,,}, ymax=0.2]
    \addplot[thick,blue] table [x=x, y=g_dist_inc, col sep=comma] {data/pucrs/rovers.csv};
    \addplot[thick,orange] table [x=x, y=g_dist_dec, col sep=comma] {data/pucrs/rovers.csv};
    \addplot[thick,teal] table [x=x, y=g_dist_ent, col sep=comma] {data/pucrs/rovers.csv};
                              
    \nextgroupplot[grid=both, every major grid/.style={gray, opacity=0.5}, yticklabels={,,}, ymax=0.2]
    \addplot[thick,blue] table [x=x, y=g_dist_inc, col sep=comma] {data/pucrs/satellite.csv};
    \addplot[thick,orange] table [x=x, y=g_dist_dec, col sep=comma] {data/pucrs/satellite.csv};
    \addplot[thick,teal] table [x=x, y=g_dist_ent, col sep=comma] {data/pucrs/satellite.csv};
                              
    \nextgroupplot[grid=both, every major grid/.style={gray, opacity=0.5}, yticklabels={,,}, ymax=0.2]
    \addplot[thick,blue] table [x=x, y=g_dist_inc, col sep=comma] {data/pucrs/blocks-world.csv};
    \addplot[thick,orange] table [x=x, y=g_dist_dec, col sep=comma] {data/pucrs/blocks-world.csv};
    \addplot[thick,teal] table [x=x, y=g_dist_ent, col sep=comma] {data/pucrs/blocks-world.csv};
                              
    \nextgroupplot[grid=both, every major grid/.style={gray, opacity=0.5}, yticklabels={,,}, ymax=0.25,xlabel={\% plan}, every axis x label/.append style={at=(ticklabel cs:-2.75)}]
    \addplot[thick,blue] table [x=x, y=g_dist_inc, col sep=comma] {data/pucrs/miconic.csv};
    \addplot[thick,orange] table [x=x, y=g_dist_dec, col sep=comma] {data/pucrs/miconic.csv};
    \addplot[thick,teal] table [x=x, y=g_dist_ent, col sep=comma] {data/pucrs/miconic.csv};
    
    \end{groupplot}

    \path (plots c1r1.north east) -- node[above=2 3]{\ref{CommonLegend}} (plots c4r1.north west);
    \end{tikzpicture}
    }

    \caption{Entropy gains and distance-to-goal measures for the automated benchmarks. }
    \label{plt:pucrs}
\end{figure*}

Two baseline communication strategies are tested alongside informative communication: increasing order, which communicates actions in plan order, and decreasing order, which communicates the actions in reversed plan order. Figure \ref{plt:pucrs} shows the average measures of the benchmark for the tested strategies. On the first row is shown the gain on entropy obtained by using the informative communication strategy rather than the others, measured as $G_{\text{inf}} = H(P_R,P^R_H) - H_{\text{inf}}(P_R,P^R_H)$. The second row instead shows the distance measure $D_G(a_i)$ between the predicates affected by a communicated action $a_i$ and the goal predicates of the instances. $D_G(a_i) = \frac{k}{|\pi_R|}$ means that, along the plan and starting counting from $a_i$, the first action directly affecting the predicates of the instance's goal is the $k$-th, eg. $D_G(a_i) = 0$ if $a_i$ has in its effects a predicate contained in the instance's goal.

The plots highlight an additional relevant property of the informative strategy, that is a consistent low distance-to-goal measure for the actions communicated earliest. This means that the informative strategy tends to communicate earlier the actions that directly affect the predicates of the goal. If we think about it, these actions are clearly informative, because communicating the goal predicates (namely, communicating the goal) of the plan clearly mostly disambiguates the goal from the others. This order of communication is similar to the decreasing strategy in its early actions, and as an average, after communicating the first 10\% of actions (that roughly correspond to one or two depending on the domain) the informative and the decreasing strategies are equally informative ($G_\text{inf} = 0$). Notice also that because of this reason the decreasing strategy is consistently more informative than the increasing strategy.

\section{User Study}

To further test our method, we performed a user study that simulates the case in which an operator encounters a robot in a corridor of a warehouse, and not knowing what it's doing asks it to verbalize its plan. The study's participants were given 12 scenarios of this type, illustrated on a Graphical User Interface (Figure~\ref{fig:user_study_gui}). In each of the scenarios there are two rooms connected by a corridor and containing 1 to 3 random objects of the same color. Objects can be circles, squares or triangles, and can be either red or blue. There are 3 possible exit doors at the corners. A recharge station is in the central corridor. The robot has 3 possible actions: moving from corridors and rooms, grabbing objects and recharging at the recharge station. In every scenario the robot had the goal of grabbing two random objects from those available, and exit the floor from a random door. It was also set to need to recharge at the recharge station in half of the scenarios. Both participants and robot had full observability on the initial state of the scenario, with robot position and available objects. However, the participants weren't informed on the belief of the robot, its goal or its plan.

\begin{table}[]
    \centering
    \resizebox{.8\linewidth}{!}{
    \begin{tabular}{|l|l|l|}
       \hline
       Increasing order & Decreasing order & Informative order\\
       \hline
       I will navigate to the central corridor. & I will exit from the bottom door. & I will grab a red circle. \\
       I will reach the warehouses doors. & I will navigate to the right corridor. &  I will grab a blue cube. \\
       I will enter the blue warehouse. &  I will reach the warehouses doors. &  I will exit from the bottom door. \\
       I will grab a blue cube. & I will grab a red circle. & I will enter the blue warehouse. \\
       I will reach the warehouses doors. & I will enter the red warehouse. & I will enter the red warehouse. \\
       I will enter the red warehouse. & I will reach the warehouses doors. & I will navigate to the central corridor. \\
       I will grab a red circle. & I will grab a blue cube. & I will navigate to the right corridor. \\
       I will reach the warehouses doors. & I will enter the blue warehouse. & I will reach the warehouses doors. \\
       I will navigate to the right corridor. & I will reach the warehouses doors. & I will reach the warehouses doors. \\
       I will exit from the bottom door. &  I will navigate to the central corridor. &  I will reach the warehouses doors. \\

       \hline
    \end{tabular}
    }
    \caption{Verbalizations of a plan from the user study, sorted by the increasing, decreasing and informative strategies.}
    \label{tbl:verbalizations}
\end{table}
At every step of a scenario participants were given a picture of the scenario and a verbalization of the robot's plan, in the form of its plan in textual form (see Table~\ref{tbl:verbalizations}). The verbalization was of increasing size at each of the scenario's steps (at the first step of size 1, second step of size 2, etc.), but, randomly for every scenarios, the verbalization could be given in one of the strategies \textit{informative, increasing} and \textit{decreasing}. Using picture and verbalization at every step the participant were asked to predict the robot's goal among a set of 3 possible goals, or to state they didn't know. Participants utilized the GUI to see the current scenario and verbalization, and to provide their prediction---for example, Figure \ref{fig:user_study_gui} shows the participants view of a scenario after the third verbalization of the robot.

We conducted the experiments with 10 participants. The participants were given a thorough introduction to the domain and the task they were asked to perform. At any time they could ask any additional clarification. In the attempt of better simulating an online interaction, the participants were told that the experiment was going to be evaluated also in term of the time required to answer the questions, and that they should have been both accurate and fast. Figure~\ref{plt:user_study} shows the average hit ratio of the participants as a function of the percentage of communicated plan. The hit ratio was measured as the number of times the correct goal was predicted divided by the number of predictions, averaged among all the participants. The average length of plans was 9.14 actions.

\begin{table}[]
    \centering
    \begin{tabular}{lr}

\begin{minipage}{.5\textwidth}
    \centering
    \includegraphics[width=.9\linewidth]{fig/user_study.png}
    \caption{\footnotesize Graphical user interface utilized during the user study. The interface shows the positions of robot, objects and doors (left). For every step of a scenario the participants could select a prediction about the robot's goal (right) from a verbalization as a written sequence of actions (bottom).}
    \label{fig:user_study_gui}
\end{minipage}

         &

\begin{minipage}{.48\textwidth}
    \centering
    \scalebox{.7}{
    \begin{tikzpicture}
    \begin{axis}[
        title={},
        xlabel={\% plan},
        ylabel={Hit ratio},
        xmin=0.1, xmax=1,
        ymin=0, ymax=1,
        xtick={0.1, 0.2, 0.3, 0.4, 0.5, 0.6, 0.7, 0.8, 0.9, 1},
        ytick={0.1, 0.2, 0.3, 0.4, 0.5, 0.6, 0.7, 0.8, 0.9, 1},
        legend pos=south east,
        ymajorgrids=true,
        grid style=dashed,
        legend entries = {Increasing, Decreasing, Informative},]
    ]
    \addplot[thick,blue] table [x=x, y=increasing, col sep=comma] {data/user_study/hit.csv};
    \addplot[thick,orange] table [x=x, y=decreasing, col sep=comma] {data/user_study/hit.csv};
    \addplot[thick,teal] table [x=x, y=informative, col sep=comma] {data/user_study/hit.csv};

    \end{axis}
    \end{tikzpicture}
    }

    \caption{\footnotesize Hit ratio of the participants as a function of percentage of communicated actions.}
    \label{plt:user_study}
\end{minipage}

         \\
    \end{tabular}
    \label{tab:my_label}
\end{table}

The results highlight the effectiveness of the informative communication strategy among the participants. After communicating 20\% of the plan, corresponding to 2 actions as an average, the hit ratio for the informative strategy was already measured as 0.85. In comparison, the decreasing strategy takes half of the plan to be as effective, while the increasing strategy almost all of the plan. The statistical significance of the informative strategy is tested against the null hypotheses $H_{\text{inc}}$: no difference with the increasing strategy ($\mu_{\text{inf}}$ = $\mu_{\text{inc}}$), and $H_{\text{dec}}$: no difference with the decreasing strategy ($\mu_{\text{inf}}$ = $\mu_{\text{dec}}$). $\mu$ is the average earliest step at which the participants answered correctly, depending on the strategy used. The corresponding found p-values are: $p(n \leq \mu_{\text{inf}}|H_{\text{inc}}) = 0.0014$ and $p(n \leq \mu_{\text{inf}}|H_{\text{dec}}) = 0.04$, showing statistical significance of the results.

\section{Discussion}
\label{sec:discussion}

Our observations are that the most informative actions to communicate are the ones that, for a planning instance, directly affect the goal predicates, therefore likely at the end of the communicated plan. Inside a planning instance communicating the goal is what transfers the highest information because a large part of the plan can be inferred as a product of it. This makes sense to us because the last actions of a plan are the most expensive to commit to. Or, from a rationality perspective inside the observer's expectation model (of which inference of plans inside the models is based on), it would be irrational to see achieving a goal and then commit to another, excluding the case where the first goal is a sub-goal of the second. After the actions affecting the goal predicates have been communicated, the actions that mostly discriminate the beliefs become the most informative. This is reflected also in tests of the user study, where the informative strategy systematically selected as first actions to communicate the robot's exit point and the gathered objects. 

Overall, the proposed informative communication works by communicating what most discriminates the robot's true state of mind among the set of other candidates, inside the estimated user's second-order theory of mind. The user study demonstrated that this strategy is effectively informative also for human participants. The proposed order of communicating plans based on informativeness, and formalized by the delivered algorithm, is mostly different from the seemingly natural way of communicating incrementally, that is also what is commonly used in the literature. Surprisingly, from our tests it results that the incremental strategy is the most inefficient among those being tested. To deepen the investigation, future work could probe whether human-human communication in task-driven scenarios follows more an informative rather than incremental strategy.

\section{Conclusion}
\label{sec:conclusion}

In this paper we proposed a strategy for the communication of robot plans that is based on the informativeness of verbalizations when contextualized inside a theory of mind. We compared the informative communication strategy with two baselines that followed increasing and decreasing plan order, finding that the informative strategy was effectively better at making users understand the robot's state using the least amount of verbalizations. 

Our results further indicated that the communication strategy most commonly proposed in the literature, namely incremental in plan order, was the most inefficient at transferring information compared to the others. This is because the informative strategy (and partly the decreasing strategy) leverage the fact that the plan is grounded in a planning instance, and communicating as earliest key parts of it such as its goal allows the explainee to infer most of the remaining parts. This was also confirmed by a user study showing an effective correlation between informativeness and the capacity of participants to predict the robot's state.

There are several interesting directions for future work based on informative communication. A first direction is about a deepened investigation on how informative communication relates to human communication in task-driven scenarios. e.g. is it more natural to use an incremental or an informative strategy? Additionally, tests in a real human-robot interaction setting could be performed. And finally, additional tests using the proposed theory of mind model on more complex scenarios could also be performed.

\bibliographystyle{splncs04}
\bibliography{bibliography}

\end{document}